\documentclass[letterpaper, 10 pt, conference]{ieeeconf}  

\IEEEoverridecommandlockouts                              

\usepackage{graphicx}
\usepackage{amsmath}
\usepackage{amssymb}
\usepackage{booktabs}
\usepackage{comment}
\usepackage{makecell}
\usepackage{tikz}
\usepackage[utf8]{inputenc}
\usetikzlibrary{calc,positioning,arrows.meta,fit,backgrounds}

\usepackage{arydshln}   
\title{\LARGE \bf
Training-Free Text-to-Image Compositional Food Generation \\ via Prompt Grafting 
}

\author{
Xinyue Pan$^{1}$, Yuhao Chen$^{2\ast}$, and Fengqing Zhu$^{1\ast}$ \\
$^{1}$Elmore Family School of Electrical and Computer Engineering, 
Purdue University, West Lafayette, IN, USA \\
$^{2}$Department of Systems Design Engineering, 
University of Waterloo, Waterloo, ON, Canada \\
}

\begin{document}

\maketitle
\thispagestyle{empty}
\pagestyle{empty}

\begin{abstract}
Real-world meal images often contain multiple food items, making reliable compositional food image generation important for applications such as image-based dietary assessment, where multi-food data augmentation is needed, and recipe visualization. However, modern text-to-image diffusion models struggle to generate accurate multi-food images due to object entanglement, where adjacent foods (e.g., rice and soup) fuse together because many foods do not have clear boundaries. To address this challenge, we introduce \textbf{Prompt Grafting (PG)}, 
a training-free framework that combines explicit spatial cues in text with implicit layout guidance during sampling. 
PG runs a two-stage process where a layout prompt first establishes distinct regions and the target prompt is grafted once layout formation stabilizes. The framework enables food entanglement control: users can specify which food items should remain separated or be intentionally mixed by editing the arrangement of layouts. Across two food datasets, our method significantly improves the presence of target objects and provides qualitative evidence of controllable separation.


\vspace{-1mm}
\end{abstract}

\section{Introduction}

Compositional food image generation is crucial for many real-world applications. In image-based dietary assessment, for example, a single eating occasion often contains multiple foods \cite{shao2021,hamid2017,luke2015}. To train a reliable food image analysis model, it often needs large amounts of diverse multi-food images. Beyond dietary assessment, the task also enables personalized recipe visualization, menu design, and food-related marketing, where clear description of multiple food items are usually required \cite{reedy2014,vermeir2020,gabrielle2012,xia2020}.

\begin{figure}[t]
    \centering
    \includegraphics[width=0.95\linewidth]{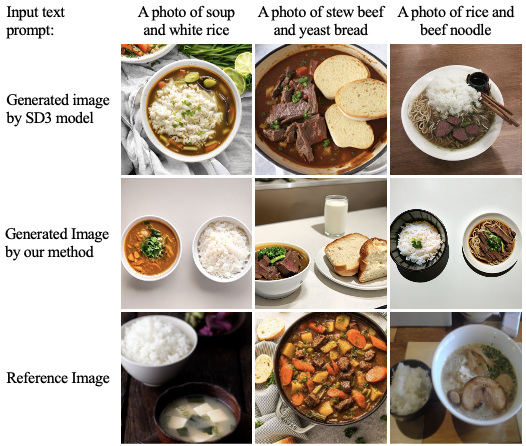}
    \caption{Example compositional food images generated by stable diffusion v3 model (SD3) and our method with corresponding reference images.}
    \label{fig:issue}
\vspace{-3mm}
\end{figure}

Although recent work on compositional image generation have achieved remarkable progress \cite{liu2022compositional,feng2022training,
chefer2023attend,rassin2023linguistic}, they still struggle with the 
object entanglement problem. This issue is common in the food domain because many food items lack clear geometric boundaries, causing adjacent foods to merge into a single, indistinct region. For example, white rice and beef noodle have soft textures and irregular shapes, as shown in Figure \ref{fig:issue}.  
Even the recent Stable Diffusion 3 (SD3) model \cite{esser2024scaling} entangles two food objects that should appear on separate plates. 

Recent text-to-image compositional generation work primarily builds upon Stable Diffusion v1 (SD1) \cite{Rombach_2022_CVPR} using only text prompts as input \cite{liu2022compositional,chefer2023attend,feng2022training,rassin2023linguistic}. However, the CLIP text encoder in SD1 \cite{radford2021learning} often attends to only one or two tokens, resulting in frequent object omissions \cite{kamath-etal-2023-text,zhang2024long}. Newer models such as SD3 \cite{esser2024scaling} and FLUX.1 \cite{labs2025flux1kontextflowmatching,flux2024} mitigate this issue through dual encoders (CLIP\cite{radford2021learning}+T5\cite{2020t5}), yet they still exhibit a strong single-object bias inherited from pretraining on LAION \cite{schuhmann2022laionb}, where most images contain one centrally placed object. As shown in Section \ref{sec:empirical_entangle}, SD3 still fails to initialize correct multi-object layouts during early denoising steps of image generation, leading to object entanglement.

To address this problem, we propose a framework built upon Prompt Grafting (PG). Before introducing PG, we acknowledge that adding simple spatial cues to the prompt (e.g., “on the lower left,” “on the top right”) can sometimes help encourage foods to appear in different regions, but this alone is unreliable, as shown in Section \ref{sec:empirical_position}. 
Therefore, we propose \textbf{Prompt Grafting (PG)} that initializes a separable layout and then places food items into layouts.  In early denoising, we condition on a layout prompt to generate distinct layouts; once layout formation stabilizes, we graft to the target prompt so that content fills those layouts. Overall, this method ensures adjacent foods remain visually separate. Crucially, our method is training-free (no finetuning beyond pretrained model) and requires no annotated layouts. Additionally, it enables food entanglement control: users simply place the items they wish to mix on the same layout. We validate the approach on two food datasets with both quantitative and qualitative ways: detection-based metrics improve markedly over SD3 baselines; we also show qualitative generalization to non-food domain.

Our contributions can be summarized as follows:
\begin{itemize}
\item To the best of our knowledge, this is the first work to address object entanglement in compositional food image generation.
\item We present empirical analysis that reveal both object‑omission and object‑entanglement issues in compositional food image generation.
\item We introduce \textbf{Prompt Grafting (PG)}, a training-free framework that disentangles overlapping objects and mitigates entanglement.
\end{itemize}

\vspace{-1mm}

\section{Related Work}
\label{sec:related}
\subsection{Food Image Generation}
High-quality food images are essential for training image analysis models in tasks like classification, segmentation, and portion estimation. Some works tackled data scarcity by synthesizing food images: Fu \textit{et al.} exposed inter-class confusion in StyleGAN-generated food and showed that isolating training to one category improves visual fidelity \cite{karras2021}. Han \textit{et al.} compared diffusion models against GANs for food synthesis, finding that diffusion approaches better handle the wide stylistic variations in cooking \cite{han2023}. Yamamoto \textit{et al.} enhanced semantic alignment by injecting CLIP embeddings into the generative process \cite{Yamamoto2022}. Li \textit{et al.}’s ChefFusion couples a Transformer-based recipe generator with a diffusion backbone to jointly produce cooking instructions and images \cite{li2024cheffusion}, while Ma \textit{et al.}’s MLA-Diff integrates ingredient–image CLIP features via an attention-fusion module for more realistic renderings \cite{ma2024}. Most recently, Yu \textit{et al.}’s CW-Food framework separates shared class-level traits from individual-instance details using a Transformer fusion layer and LoRA fine-tuning, enabling diverse food synthesis from text labels alone \cite{yu2025}. Despite these advances in single food and recipe-guided generation, none directly address the unique challenges of composing multiple distinct food items into one coherent form.  

\subsection{Compositional Image Generation}
Compositional image generation aims to generate an image containing multiple, specified objects. Two core challenges arise: (1) \textbf{attribute binding}, ensuring attributes (e.g., color, size) correctly attach to their objects; and (2) \textbf{object existence}, preventing the omission of required items. There are some related works that have used Stable Diffusion v1 model (SD) in this task by manipulating attention maps during the denoising process. Liu \textit{et al.} dispatch separate diffusion streams for different prompt components, improving token–object alignment \cite{liu2022compositional}. Feng \textit{et al.} proposed \textbf{Structured Diffusion}, which parses prompts into paired entities (e.g., “red car,” “white house”) 
and applies structured guidance to enforce correct attribute–object pairings \cite{feng2022training}. Chefer \textit{et al.} introduce an \textbf{Attend and Excite} pipeline that increases weights on low‑attention tokens to reduce object omission \cite{chefer2023attend}. Rassin \textit{et al.} proposed \textbf{Synegn}, which leverages syntactic parsing and a custom loss to align related words and prevent incorrect bindings \cite{rassin2023linguistic}. Despite their success, all of these methods are built on SD’s CLIP text encoder, which is limited to simple prompts, and none of them address the object entanglement problem in image generation.

Several methods introduce additional modules to achieve compositional control, such as LayoutDiffusion \cite{zheng2023layoutdiffusion}, GLIGEN \cite{li2023gligen}, and Instance Diffusion \cite{wang2024instancediffusion}. While effective, these approaches require architectural extensions, more GPU resources, and additional inputs beyond a text prompt such as bounding box. In contrast, our work focuses on an training-free strategy, uses text prompts as the only input, and avoids modifying the diffusion architecture.

\vspace{-1mm}

\section{Empirical Studies}
\label{sec:empirical}
We analyze how diffusion models handle compositional food prompts by inspecting token attention for selected text tokens to answer questions whether stronger text encoder solves the problem, what causes object entanglement problem and whether spatial cues in text prompt help solve object entanglement problem. 
\subsection{Object Omission in Stable Diffusion v1 Model}

\begin{figure}
    \centering
    \includegraphics[width=0.95\linewidth]{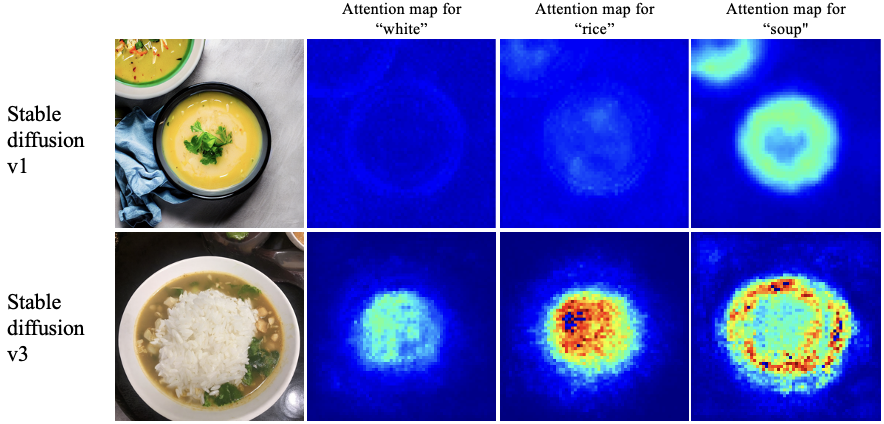}
    \caption{Generated image from stable diffusion v1 and stable diffusion v3 model using text prompt: A photo of white rice and soup}
    \label{fig:emp_study1}
    \vspace{-2mm}
\end{figure}

Most prior training-free compositional image generation models (e.g., Structured Diffusion\cite{feng2022training}, Attend \& Excite \cite{chefer2023attend}, Syngen\cite{rassin2023linguistic}) are developed and evaluated on Stable Diffusion v1 (SD1)\cite{Rombach_2022_CVPR}, whose CLIP text encoder \cite{radford2021learning} exhibits obvious object-omission problem. As shown in Figure~\ref{fig:emp_study1}, when prompting SD1 with “A photo of white rice and soup,” the generated image contains only the soup, entirely omitting the rice. The corresponding cross-attention maps show that the “soup” token monopolizes nearly all attention weight, while the “white” and “rice” tokens receive almost zero weight. This phenomenon traces to CLIP’s pretraining on predominantly single-object data, which biases the model toward focusing on a single salient token in multi-object prompts. In contrast, Stable Diffusion v3 (SD3) leverages a dual‑encoder, including a T5 encoder that handles longer, syntax‑rich text, to distribute attention more evenly across multiple tokens. As shown in Figures \ref{fig:emp_study1}, 
SD3 assigns non‑zero attention weights to all specified tokens, resulting in the inclusion of each requested object in the output. 

While SD1 highlights the severity of the omission problem, SD3’s improved text encoder largely resolves omission and reveals object entanglement problem, where multiple foods are merged together. Because SD3 resolves object omission issue, it provides motivation for us to develop our method on SD3 rather than SD1.

\subsection{Object Entanglement in Stable Diffusion v3}
\label{sec:empirical_entangle}
\begin{figure}
    \centering
    \includegraphics[width=0.95\linewidth]{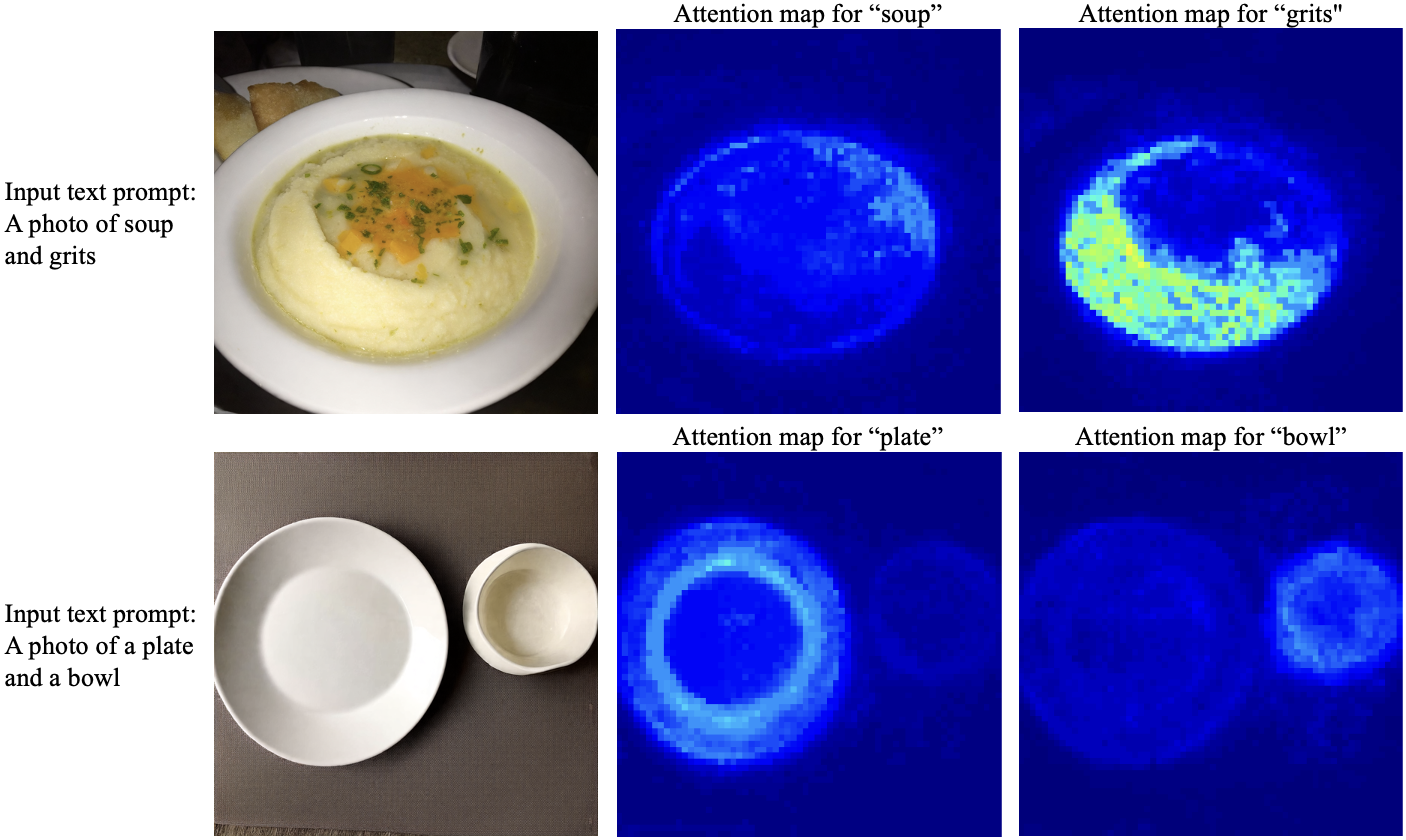}
    \caption{Generated image from stable diffusion v3 with input text prompt from ``soup and grits" and ``plate and bowl"}
    \label{fig:emp_study4}
    \vspace{-2mm}
\end{figure}
\begin{figure}
    \centering
    \includegraphics[width=0.95\linewidth]{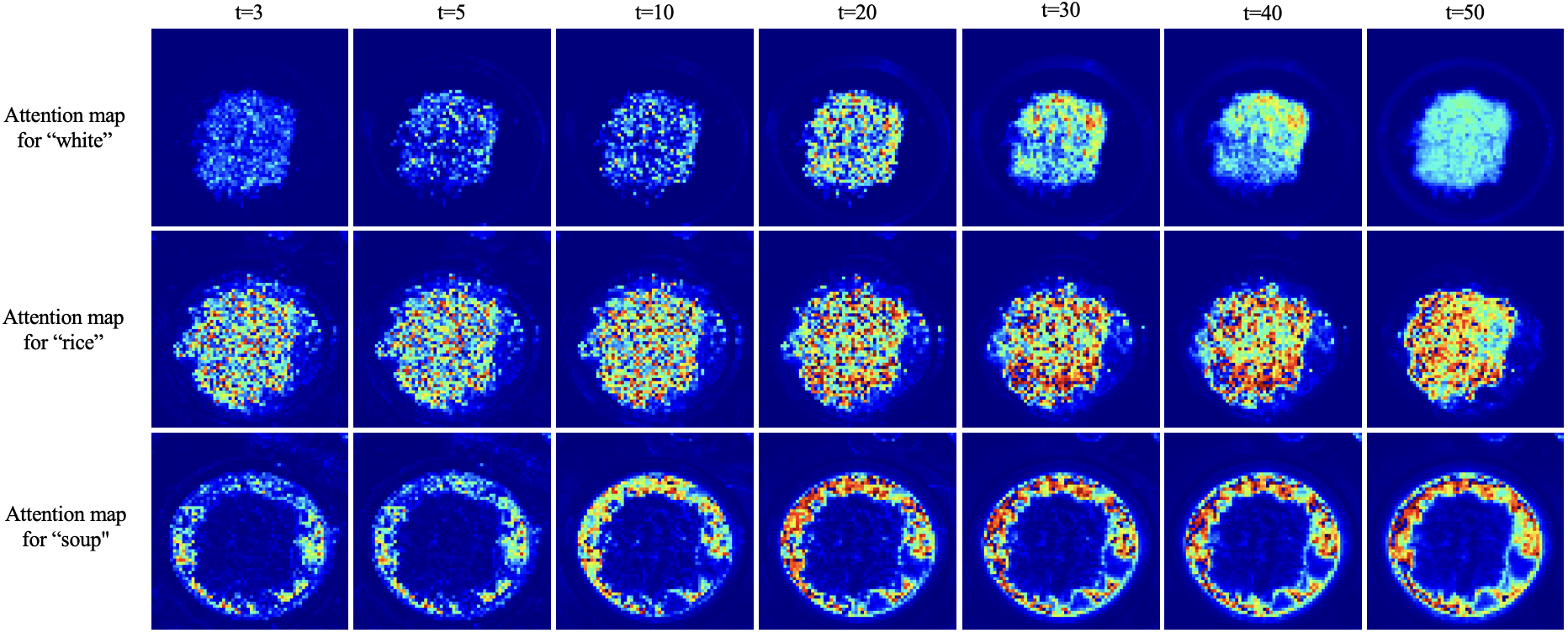}
    \caption{Attention map for selected text tokens in the first 50 inference steps for Stable Diffusion v3 model given 100 total inference steps with text prompt input: ``A photo of white rice and soup" Once layout formation stabilizes, later steps only refine details within the layout.}
    \label{fig:emp_study6}
    \vspace{-2mm}
\end{figure}
\begin{figure}
    \centering
    \includegraphics[width=0.95\linewidth]{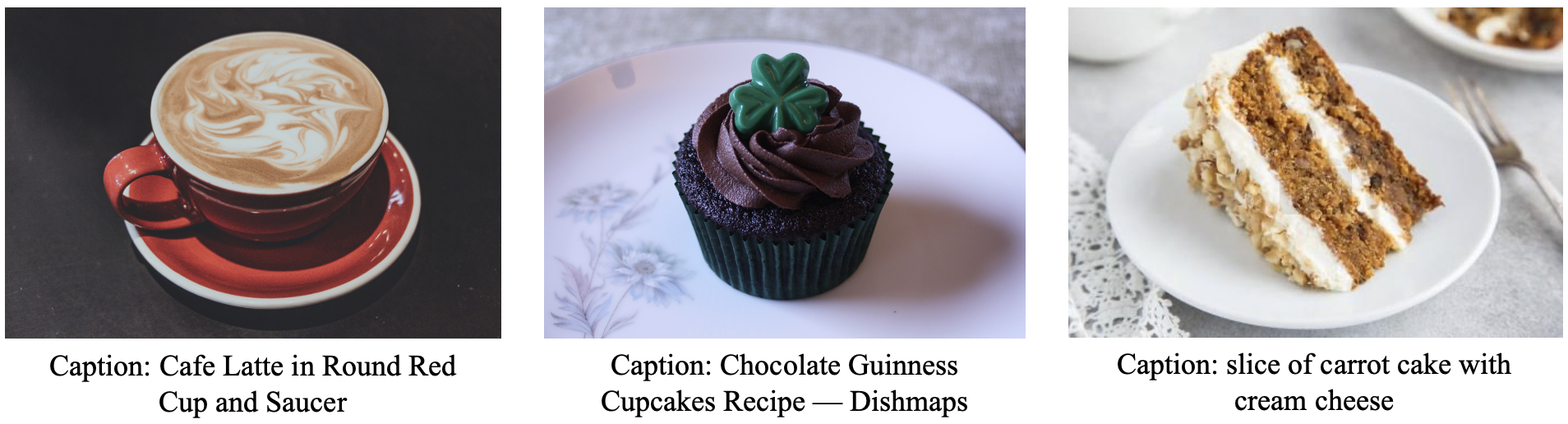}
    \caption{Examples of image-caption data stable diffusion model pretrained on}
    \label{fig:laion-2b}
\vspace{-2mm}
\end{figure}
Although SD3 rarely omits requested items, it still suffers from object entanglement, with distinct foods merging into a single, indistinct region. For example, in Figure \ref{fig:emp_study1}, “white rice” and “soup” fuse despite both tokens receiving non-zero attention. 
Unlike objects with clear contours (e.g.\ plates versus bowls), food items often lack well-defined edges, making them especially prone to fusion (Figure \ref{fig:emp_study4}). We trace this behavior to SD3’s early denoising phase, which establishes a unified coarse layout before adding fine details. As Yi et al. \cite{yi2024towards} demonstrate, latent diffusion models first generate a rough spatial structure in the initial steps; if that layout merges adjacent regions, later refinements cannot separate them. This can also be shown in Figure \ref{fig:emp_study6}, where the attention maps at different inference steps for text token ``white," ``rice" and ``soup" are shown. The rice and soup layout have been formed very early in inference steps, and in the later inference steps,  attention shifts toward fine-grained details such as textures and edges (warmer colors in Figure~\ref{fig:emp_study6} highlight this refinement), while the overall layout remains fixed. This factor is likely carried over from SD3's pretraining on LAION\cite{schuhmann2022laionb} subsets, where subjects are usually centered, resulting in generated objects bias toward the image center (Figure \ref{fig:laion-2b}). 

\subsection{Incorporation of Spatial Cues in Text Prompt}
\label{sec:empirical_position}
\begin{figure*}
    \centering
    \includegraphics[width=0.9\linewidth]{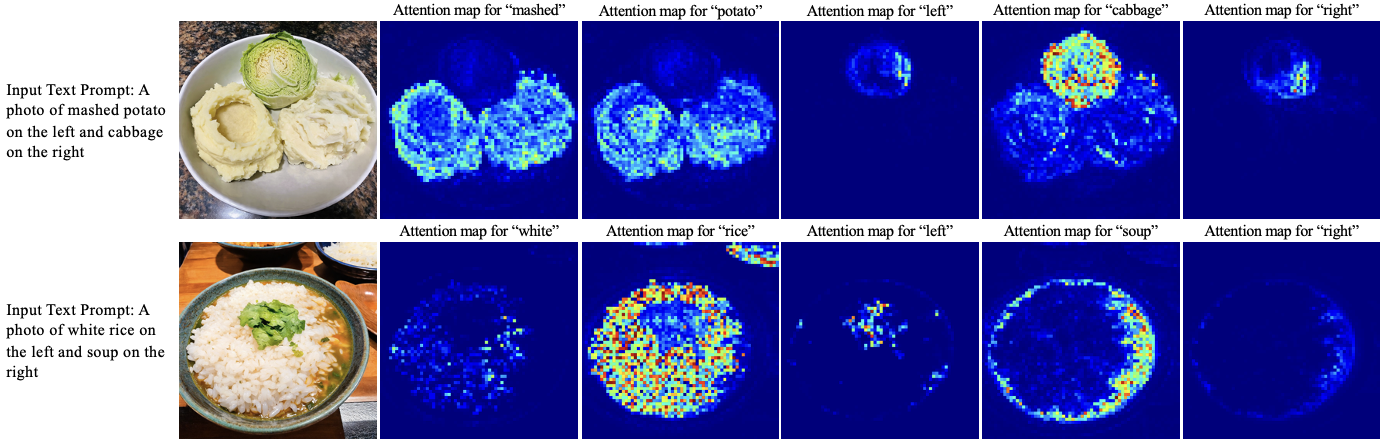}
    \caption{Generated image and selected attention maps from Stable Diffusion v3 model using text prompt with spatial cues}
    \label{fig:emp_study8}
\vspace{-2mm}
\end{figure*}

We investigate whether adding spatial cues to the prompt can reliably guide SD3’s layout formation. As shown in Figure \ref{fig:emp_study8}, a simple prompt such as ``A photo of mashed potato on the left and cabbage on the right'' can separate the two items, but a similar prompt like ``A photo of rice on the left and soup on the right'' still produces a single fused object, and the cross-attention maps show that the ``left'' and ``right'' tokens are not consistently grounded in the correct regions. These results reveal two limitations: first, the text encoder does not reliably interpret spatial relationships, causing objects placed into wrong region, i.e. cabbage placed in the center instead of right; second, even when the model roughly places items in correct region, it has no interpretation of how far apart the objects should be, leading to foods that blend together. Therefore, spatial cues alone are not enough to prevent object entanglement. These observations motivate our approach of combining explicit spatial cues with an early layout initialization interruption (Prompt Grafting) so that SD3 forms clearly separated regions before filling them with fine-grained food details.

\vspace{-1mm}

\section{Method}
\label{sec:method}
\begin{figure*}
    \centering \includegraphics[width=0.9\linewidth]{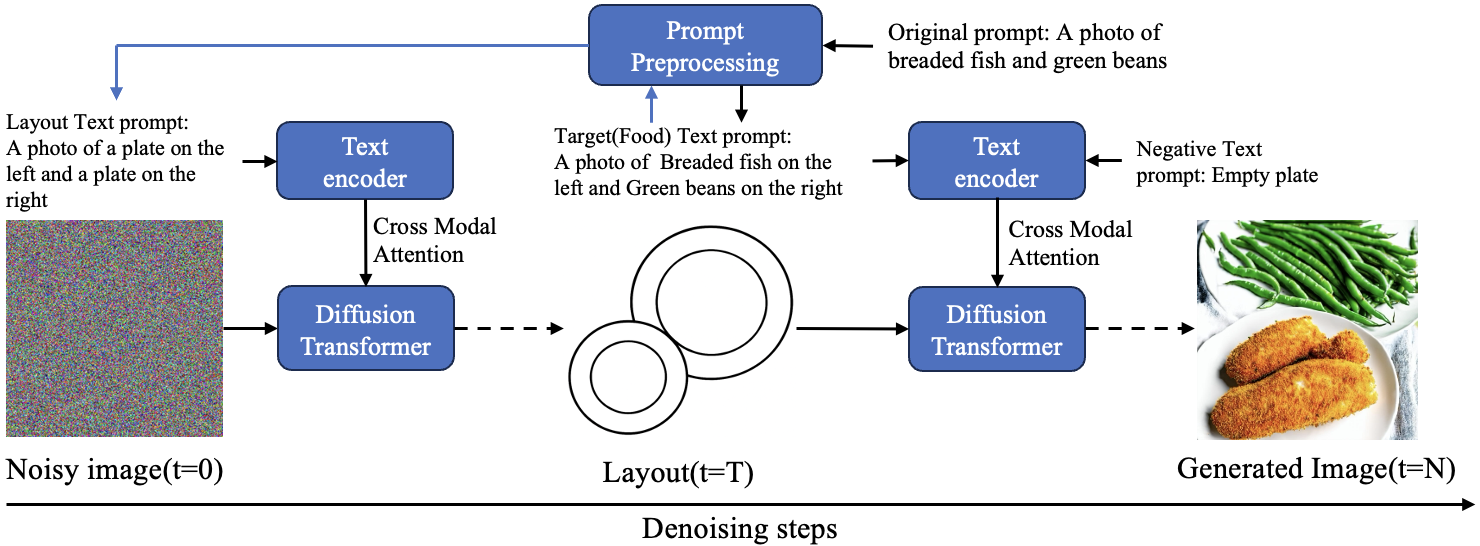}
    \caption{Overview of our method: We propose prompt grafting by using the layout prompt in the first T denoising steps and use target prompts in the later timesteps, which are both generated by Large Language Model(LLM). }
    \label{fig:pipeline}
\vspace{-3mm}
\end{figure*}

Figure~\ref{fig:pipeline} provides an overview of our training-free method (no finetuning beyond pretrained model), and it requires no modification to the SD3 architecture. Our approach is based on an empirical observation (Section~\ref{sec:empirical}): the early denoising steps determine the global layout, while later steps fill in fine-grained details in each layout. Object entanglement occurs when SD3 initializes an incorrect layout. Prompt Grafting (PG) addresses this by separating the process into two stages. For the first $T$ steps, we condition the model on a layout prompt (e.g., “A photo of a plate on the left and a plate on the right”) so that SD3 forms distinct regions. After the layout formation stabilizes, we graft to the target prompt, which fills each region with the intended food items. This works because SD3 can separate general objects(e.g, plates, bowls, trays) since they have clear shapes and boundaries, whereas many foods do not have clear boundaries and are more likely to blend. 
Additionally, PG offers a lightweight text-only alternative to layout-based diffusion methods. Unlike approaches such as LayoutDiffusion\cite{zheng2023layoutdiffusion}, GLIGEN\cite{li2023gligen}, or Instance Diffusion\cite{wang2024instancediffusion}, our method requires no bounding boxes, extra modules, or additional training, making it practical when only text prompts are available.

\subsection{Prompt Grafting (PG)}
\label{sec:pg}
\noindent \textbf{Data Preprocessing -- Spatial Cues (SC) in Prompts:} We add simple spatial cues (e.g., left, right, center) to the text prompt to indicate coarse regions where each food might appear, i.e.
“A photo of sushi on the left and tempura on the right.”
However, we find that spatial cues alone are insufficient for reliable separation. As discussed in Section \ref{sec:empirical_position}, the text encoder often misinterprets positional information, and the model cannot infer how far apart the foods should be. As a result, foods may still end up on the same plate or partially fused. This limitation motivates us to interrupt layout formation in early denoising steps.

\noindent\textbf{Layout Interruption:} As discussed in Section~\ref{sec:empirical_entangle}, object entanglement in SD3 occurs when its early denoising steps produce a merged spatial layout that later refinements cannot separate. To prevent this, we interrupt the layout initialization: for the first $T$ timesteps (a hyperparameter discussed in grafting timestep determination later), we condition the model on a layout prompt that enforces distinct regions; then, for the remaining steps, we graft to the target (food) prompt, allowing the model to populate each region with fine-grained details. The layout prompt is constructed by replacing food objects in the target prompt with generic container words such as ``plate'' or ``bowl'' depending on the food category. The exact container type is not critical: any easily distinguishable receptacle (e.g., plate, bowl, tray) works as long as it provides clear shape boundaries. 

Therefore, PG can be formulated as follows. Recall that SD3 sampling 
solves the ordinary differential equation (ODE) to infer the velocity field of noise at each timestep of the denoising process in SD3:

\vspace{-1mm}
\begin{equation}
  \frac{dx}{dt} \;=\; f_\theta\bigl(x(t),\,t;\,c\bigr),
  \quad x(T)\sim\mathcal{N}(0,I),
\vspace{-1mm}
\end{equation}
where \(c\) is the text‐conditioning embedding. In our approach, we make \(c\) a time‐dependent prompt,
where \(c_{\mathrm{layout}}\) denotes the embedding of the layout prompt (e.g., “a photo of a plate on the left and a plate on the right”), and \(c_{\mathrm{target}}\) indicates the embedding of the target prompt (e.g., “a photo of soup on the left and rice on the right”). 
Thus, the continuous‐time sampling becomes:
\vspace{-1mm}
\begin{equation}
  \frac{dx}{dt} = 
    \begin{cases}
      f_\theta\bigl(x,\,t;\,c_{\mathrm{layout}}\bigr), & t \le T,\\
      f_\theta\bigl(x,\,t;\,c_{\mathrm{target}}\bigr),   & t > T.
    \end{cases}
\vspace{-1mm}
\end{equation}

At each step \(t\), the latent \(x_{t-1}\) is obtained from \(x_t\) via:
\vspace{-1mm}
\begin{equation}
  x_{t-1}
    = x_t + \gamma\,f_\theta\bigl(x_t,\,t;\,c(t)\bigr),
\vspace{-1mm}
\end{equation}
where \(\gamma\) is the time increment and \(c(t)\) transitions from the layout prompt to the target prompt at timestep \(T\).

In addition to grafting between layout and target prompts, we employ classifier‑free guidance with a negative prompt ``Empty plate/bowl/tray" to discourage all food items collapsing onto a single plate during the final refinement steps, with formulation as following:
\vspace{-1mm}
\begin{equation}
\begin{split}
\hat{\varepsilon}_\theta(x_t, t)
= \varepsilon_\theta(x_t, t)
+ \omega\Bigl(\varepsilon_\theta(x_t, t \mid c_{\mathrm{target}}) \\
- \varepsilon_\theta(x_t, t \mid c_{\mathrm{neg}})\Bigr),
\end{split}
\vspace{-1mm}
\end{equation}

\(x_t\) denotes the latent image at denoising step \(t\), \(\varepsilon_\theta(\cdot)\) is the unconditional noise predictor, \(\varepsilon_\theta(\cdot \mid c_{\mathrm{target}}) \) and \(\varepsilon_\theta(\cdot \mid c_{\mathrm{neg}})\) are the noise predictors conditioned on the target prompt and negative prompt
, respectively, and \(\omega\) is the guidance scale. 

\noindent \textbf{Grafting Timestep Determination: }Selecting the grafting timestep \(T\) is critical: switching too early yields under-defined layouts; switching too late over-generates plate/layout formation. 
We assume that the quality of the coarse layout depends on the stability of semantic structure in the latent space. 
We monitor the CLIP image–text similarity between the current decoded latent and the layout prompt as a proxy for semantic stability. When this similarity plateaus, indicating that layout formation has stabilized, we dynamically switch from the layout prompt to the target food prompt. At diffusion step \(t\), let
\vspace{-1mm}
\begin{equation}
S_{\mathrm{lay}}(t) \;=\; \mathrm{sim}_{\mathrm{CLIP}}\!\big(f(x_t),\, p_{\mathrm{layout}}\big),
\vspace{-1mm}
\end{equation}
where \(x_t\) is the current latent, \(f(\cdot)\) denotes decoded latent, and \(p_{\mathrm{layout}}\) is the layout prompt. We monitor a \(k\)-step difference and trigger grafting at the first \(t\) in a safe window \([t_{\min}, t_{\max}]\) such that
\vspace{-1mm}
\begin{equation}
\Delta_k(t) \;=\; S_{\mathrm{lay}}(t) - S_{\mathrm{lay}}(t-k) \;\le\; \varepsilon,
\qquad t \in [t_{\min},\, t_{\max}].
\vspace{-1mm}
\end{equation}
In practice we use small \(k\) (e.g., \(k=2\)) and a tight tolerance (\(\varepsilon=0.002\)), check every step, and bound the search with \(t_{\min}=0.02\,T_{\text{total}}\) and \(t_{\max}=0.20\,T_{\text{total}}\) to avoid grafting too early or too late; empirically, exceeding \(20\%\) of the denoising steps often leads to plate-only content in the generated image. 

\vspace{-1mm}

\section{Experiments}
\label{sec:experiments}

\begin{figure*}[t]
   \centering
   \includegraphics[width=1\linewidth]{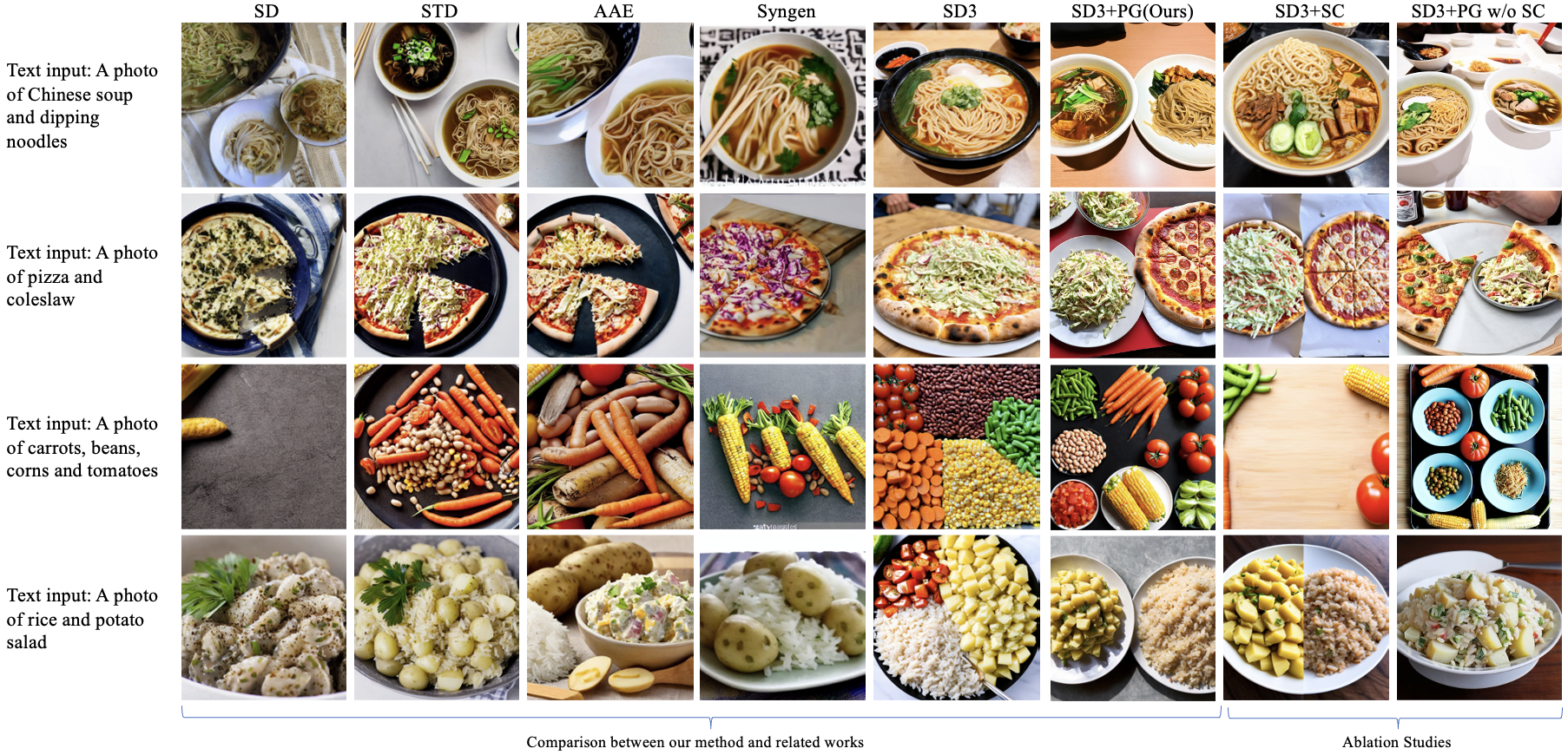}
   \caption{\textbf{Comparison of our method with related works:} Our approach (PG) yields clearly separated food items by first establishing distinct plate regions via layout interruption, then grafting the target prompt. Example text prompt with SC: \textbf{Rice+Potato salad:} Target prompt: ``A photo of rice on the left and potato on the right " Layout prompt: ``A photo of a plate on the left and a plate on the right" }
   \label{fig:qual_result}
\vspace{-3mm}
\end{figure*}

We validate our approach on two food datasets, VFN \cite{mao2020,he2023long} and UEC-256 \cite{kawano14c}, using YOLOv11 detection metrics (precision, recall, F1), BLIP in vision question answering for object existence rate and Fréchet Inception Distance (FID) as quantitative measures. In addition to standard baselines, we present ablation studies to evaluate the effect of incorporating spatial cues (SC) in prompt grafting, and we include qualitative examples to illustrate object entanglement problem visually. Across both datasets, our combined method obviously reduces entanglement issue and boosts object detection rates, and we further demonstrate its applicability to non‐food domain.

\subsection{Datasets}
Since our evaluation metrics include the detection on whether a food object exist or not in a generated image, we select two food datasets that include bounding‐box annotations for image generation. For both datasets, we only test on those containing multiple foods.

\noindent\textbf{VFN\cite{mao2020,he2023long}:}  
The VFN dataset contains 37k images, of which approximately 2.2k feature multiple food items. Each image is annotated with a food category label and corresponding bounding‐box coordinates. The categories are drawn from commonly consumed foods in the What We Eat In America (WWEIA) database\cite{eicher2017}. 

\noindent\textbf{UEC‑256\cite{kawano14c}:}  
UEC‑256 comprises 22k images, including around 1.7k that contain multiple foods. Each image is annotated with Japanese food categories and bounding boxes. 

\subsection{Evaluation Metrics}
To evaluate whether our generated images include all prompt-specified food items and maintain high visual quality, we employ three complementary metrics. First, we train \textbf{YOLOv11\cite{yolo2023}}, which is an object detection model, on each food dataset and report its precision, recall, and F1-score as an approximate measure of object presence (noting that small objects or rare classes may sometimes be missed). Second, we leverage \textbf{BLIP\cite{li2022blip}} in a VQA setting by asking “Yes or no: Is there a [category] in the image?” to verify object existence from a vision language perspective, utilizing BLIP’s large amount of pretrained real-world knowledge. Finally, we compute the \textbf{Fréchet Inception Distance (FID)} between generated images and real images to compute overall fidelity. Together, these metrics capture both the correctness of generating individual food items and quality of the generated images.

\subsection{Experiment Setup}

We utilize the Stable Diffusion v3 (SD3) model with the pretrained checkpoint ``stabilityai/stable-diffusion-3-medium-diffusers" without further training on datasets.  

To evaluate our approach against existing inference phase methods, we compare to \textbf{Stable Diffusion (SD) \cite{Rombach_2022_CVPR}}, \textbf{Structured Diffusion (STD)\cite{liu2022compositional}},  \textbf{Attend \& Excite (AAE)\cite{chefer2023attend}}, \textbf{Syngen\cite{rassin2023linguistic}}, \textbf{FLUX.1 \cite{labs2025flux1kontextflowmatching, flux2024}} and \textbf{Stable Diffusion v3 (SD3) \cite{esser2024scaling}}.
All models generate food images under equivalent parameters with 100 inference steps, guidance scale \( \omega = 12 \) because early steps are used for layout, fewer remain for the target prompt, which motivates us to use a moderately higher guidance scale to ensure strong conditioning within the shorter refinement phase.

\begin{table}[t]
\begin{center}
\footnotesize
\caption{YOLOv11 and BLIP detection and FID score results on generated images in VFN dataset
} 
\begin{tabular}{cccccc}
 \toprule 
 \makecell{Generation\\ Method} &\makecell{Precision \\ \(\uparrow\)} &\makecell{Recall \\ \(\uparrow\)} &\makecell{F-1 score \\ \(\uparrow\)} &\makecell{BLIP\\ exist\\ score \(\uparrow\)} &\makecell{FID\\score \\ \(\downarrow\)} 
 \\
 \midrule 
 \textbf{\makecell{Real images\\ (Ref \\ Value)}} & 0.749 & 0.779 & 0.763 & 97.9 &\\
 \midrule
 \textbf{\makecell{SD\cite{Rombach_2022_CVPR}}} & 0.252 & 0.165 & 0.199  & 82.1 & 40.5\\
 \textbf{\makecell{STD \cite{liu2022compositional}}} & 0.469 & 0.336 & 0.392 & 90.5 & 45.2 \\
 \textbf{\makecell{AAE \cite{chefer2023attend}}} & 0.509 & 0.387 & 0.439 & 95.7 
 & 39.6 \\
 \textbf{\makecell{Syngen\cite{rassin2023linguistic}}} & 0.484 & 0.312 & 0.379 & 95.8 & 40.6 \\

 \textbf{\makecell{FLUX.1 \cite{labs2025flux1kontextflowmatching, flux2024}}} & 0.52 & 0.442 & 0.478 & 99.4 & \textbf{40.5}\\
  \textbf{\makecell{SD3\cite{esser2024scaling}}} & 0.533 & 0.454 & 0.490 & 99.4 & \textbf{40.5}\\
  \midrule
  \multicolumn{6}{c}{\textbf{Our Work}}\\
  \midrule
  \textbf{\makecell{SD3\cite{esser2024scaling} \\ +PG w/o SC }} & 0.513 & 0.487 & 0.5 & 99.5 & 43.7\\
  \textbf{\makecell{SD3\cite{esser2024scaling}+SC}} & 0.54 & 0.48 & 0.508 & 99.2 & 47.8\\
  \textbf{\makecell{SD3\cite{esser2024scaling}+PG }} & \textbf{0.558} & \textbf{0.518} &\textbf{0.537} & \textbf{99.6} & 49.0\\
  \bottomrule
\end{tabular}
\label{tab:yolo_detect1}
\end{center}
\vspace{-5mm}
\end{table}

\begin{table}[t]
\begin{center}
\footnotesize
\caption{YOLOv11 and BLIP detection and FID score results on generated images in UEC-256 dataset
} 
\begin{tabular}{cccccc}
 \toprule 
 \makecell{Generation\\ Method} &\makecell{Precision \\ \(\uparrow\)} &\makecell{Recall \\ \(\uparrow\)} &\makecell{F-1 score \\ \(\uparrow\)} &\makecell{BLIP\\ exist\\ score \(\uparrow\)} &\makecell{FID\\score \\ \(\downarrow\)} 
 \\
 \midrule 
 \textbf{\makecell{Real images\\ (Ref \\ Value)}} & 0.543 & 0.492 & 0.516 & 97.2 &\\
 \midrule
 \textbf{\makecell{SD\cite{Rombach_2022_CVPR}}} & 0.057 & 0.008 & 0.014  & 91.3 & 71.9\\
 \textbf{\makecell{STD\cite{liu2022compositional}}} & 0.095 & 0.03 & 0.046 & 95.4 & 86.2 \\
 \textbf{\makecell{AAE\cite{chefer2023attend}}} & 0.112 & 0.028 & 0.045 & 97.2 & 73.8 \\
 \textbf{\makecell{Syngen\cite{rassin2023linguistic}}} & 0.173 & 0.062 & 0.091 & 97.5 & 82.4 \\
  \textbf{\makecell{FLUX.1 \cite{labs2025flux1kontextflowmatching, flux2024}}} & 0.134 & 0.05 & 0.073 & 99.5 & 69.4\\
  \textbf{\makecell{SD3\cite{esser2024scaling}}} & 0.113 & 0.038 & 0.056 & 99.5 & 70.6\\
  \midrule
  \multicolumn{6}{c}{\textbf{Our Work}}\\
  \midrule
  \textbf{\makecell{SD3\cite{esser2024scaling} \\ +PG w/o SC }} & 0.178 & 0.127 & 0.149 & \textbf{99.7} & \textbf{60.8}\\
  \textbf{\makecell{SD3\cite{esser2024scaling}+SC}} & 0.13 & 0.059 & 0.081 & 99.5 & 64.4\\
  \textbf{\makecell{SD3\cite{esser2024scaling}+PG }} & \textbf{0.216} & \textbf{0.134} &\textbf{0.165} &\textbf{99.7} & 65.0\\
  \bottomrule
\end{tabular}
\label{tab:yolo_detect3}
\end{center}
 \vspace{-5mm}
\end{table}

\begin{table}[t]
\begin{center}
\footnotesize
\caption{Ablation studies on timestep to graft the text prompt on VFN dataset
} 
\begin{tabular}{ccccccc}
 \toprule
 \makecell{Total\\ inference\\ steps} &\makecell{Layout\\ steps}  & \makecell{Precision \\ \(\uparrow\)} & \makecell{Recall \\ \(\uparrow\)} &\makecell{F-1\\ score \\ \(\uparrow\)} &\makecell{BLIP\\exist\\score \(\uparrow\)} &\makecell{FID\\ score \\ \(\downarrow\)}
 \\
 \midrule
 \textbf{\makecell{Real \\images\\ (Ref\\ Value)}}  & & 0.749 & 0.779 & 0.773 & 97.9 &\\
 \midrule
 \textbf{\makecell{100, \\ SC only }} & 0 & 0.54 & 0.48 & 0.501 & 99.4 & 47.8\\
 \midrule
 \textbf{100} & 3 & 0.534 & 0.491 &0.512 & 99.5 & 45.5\\
 \textbf{100} & 5 & 0.531 & 0.504 &0.517 & 99.3 & 55.3\\
 \textbf{100} & 7 & 0.501 & 0.478 &0.489 & 99.1 & 64.4\\
 \textbf{100} & 10 & 0.464 & 0.446 & 0.505 & 98.6 & 68.6\\
 \textbf{100} & \makecell{Varied \\ T} & \textbf{0.558} & \textbf{0.518} &\textbf{0.537} & \textbf{99.6} & 49.0\\
 \bottomrule
\end{tabular}
\label{tab:yolo_detect_ablation5}
\end{center}
 \vspace{-5mm}
\end{table}

\begin{table}[t]
\begin{center}
\footnotesize
\caption{Ablation studies on timestep to graft the text prompt on UEC-256 dataset
} 
\begin{tabular}{ccccccc}
 \toprule
 \makecell{Total\\ inference\\ steps} &\makecell{Layout\\ steps}  & \makecell{Precision \\ \(\uparrow\)} & \makecell{Recall \\ \(\uparrow\)} &\makecell{F-1\\ score \\ \(\uparrow\)} &\makecell{BLIP\\exist\\score \(\uparrow\)} &\makecell{FID\\ score \\ \(\downarrow\)}
 \\
 \midrule
 \textbf{\makecell{Real \\ images\\ (Ref\\ Value)}}  & & 0.543 & 0.492 & 0.516 & 97.2 & \\
 \midrule
 \textbf{\makecell{100, \\ SC only }} & 0 & 0.13 & 0.059 & 0.081 & 99.5 & 64.4\\
 \midrule
 \textbf{100} & 3 & 0.189 & 0.128 & 0.153 & 99.6 & 62.1\\
 \textbf{100} & 5 & 0.192 & 0.125 & 0.151 & 99.5 & 69.8\\
 \textbf{100} & 7 & 0.214 & 0.115 & 0.150 & 99.2 & 76.5 \\
 \textbf{100} & 10 & 0.2 & 0.095 & 0.128 & 99.3 & 82.4\\
 \textbf{100} &\makecell{Varied \\ T} & \textbf{0.216} & \textbf{0.134} & \textbf{0.165} &\textbf{99.7} & 65.0\\
 \bottomrule
\end{tabular}
\label{tab:yolo_detect_ablation6}
\end{center}
 \vspace{-5mm}
\end{table}

\subsection{Quantitative Results}
\label{sec:exp_result}
\noindent\textbf{Quantitative Results:}  
Tables \ref{tab:yolo_detect1} and \ref{tab:yolo_detect3} report our quantitative results on VFN and UEC-256 dataset, respectively, using YOLOv11 detection precision, recall, F1, BLIP VQA object existence rate, and FID score by comparing our proposed methods with related works. “Real images” serve as a reference value for these metrics.  

On VFN (Table \ref{tab:yolo_detect1}), vanilla SD3 already outperforms all SD based baselines. The ablation studies show that adding only spatial cues (SC-only) or applying only layout interruption (PG without SC) provides a small improvement in recall but does not reliably prevent entanglement. The full PG method, which combines layout interruption with spatial cues, achieves the best overall performance with an F1 of 0.537. This confirms that both components are necessary.

On UEC-256 (Table \ref{tab:yolo_detect3}), all methods, including ours, show lower YOLOv11 detection scores. This is largely due to the stable diffusion model’s difficulty in generating specialized Japanese foods (e.g.\ natto) that are underrepresented in pretraining. Additionally, the YOLOv11 model has some difficulty in recognizing UEC-256 objects due to high inter-class similarity issue in this dataset. Accordingly, we interpret the UEC-256 YOLOv11 object detection rate with caution. Even so, our method achieves the highest F1 score (0.165) and BLIP object existence score (99.7\%).

Overall, full PG (layout interruption and spatial cues) consistently improves object recall and existence compared to all SD based methods (Structured Diffusion, Attend \& Excite, Syngen), FLUX.1 and vanilla SD3. SC provides explicit spatial guidance, while layout interruption prevents early layout fusion, together addressing the entanglement issue. 
The increase in FID compared to SD3 and FLUX.1 (e.g., 49.0 vs. 40.5 on VFN) is expected because by interrupting the early layout and enforcing separate plates, PG reduces the diversity and complexity of backgrounds. 

\noindent\textbf{Ablation studies on layout steps in PG:} We evaluated Prompt Grafting with fixed layout inference steps \(T\in\{3,5,7,10\}\) given 100 total inference steps versus dynamic grafting timestep. On VFN (Table \ref{tab:yolo_detect_ablation5}), the highest F1 score and BLIP object existence score occur at dynamic grafting timestep, outperforming all fixed‐step settings. Similarly, on UEC‐256 (Table \ref{tab:yolo_detect_ablation6}), dynamic grafting yields the top F1 score and BLIP object existence score. These results confirm that dynamically choosing grafting timestep improves generated image quality. 
\vspace{-0.5mm}


\subsection{Qualitative Results}  
Figure \ref{fig:qual_result} compares our proposed method with related work. 
Images generated by Stable Diffusion based baselines (SD, Structured Diffusion, Syngen, Attend \& Excite) either drop one of the requested foods or, when both appear, merge them into a single, indistinct food item. In “pizza and coleslaw,” for instance, SD and STD omit one item, while AAE and Syngen fuses them together into a single food item. Even vanilla SD3, with its improved CLIP + T5 text encoding, still collapses adjacent food items into a center‐biased layout, i.e. “pizza and coleslaw” ends up a single item like ``coleslaw pizza", showing that better text encoder alone does not solve entanglement issue.


Our ablation results show how the two components inside PG address these failure cases in complementary ways. When using only spatial cues (SC only), the spatial cues (“on the left,” “on the right,” etc.) help push the foods toward different regions, and in some cases (e.g., “rice and potato salad”) the items appear more separated than with vanilla SD3. However, because the text encoder does not fully understand spatial relations or distance, SC only cannot reliably prevent the foods from sharing one plate or partially merging. When using only layout interruption (PG without SC), the model is forced to allocate multiple receptacle regions, which reduces early entanglement, but without any directional hints it may still place multiple foods into a single region or leave other regions empty (e.g., “pizza and coleslaw”). In contrast, the full PG method, which combines layout interruption with spatial cues, consistently produces two separate plates with the correct food in each plate. This demonstrates that spatial cues provide coarse directional guidance, while layout interruption enforces distinct regions; both are needed together to reliably avoid entanglement.

\subsection{Controllable Food Image Generation}

\begin{figure}[t]
   \centering
   \includegraphics[width=0.95\linewidth]{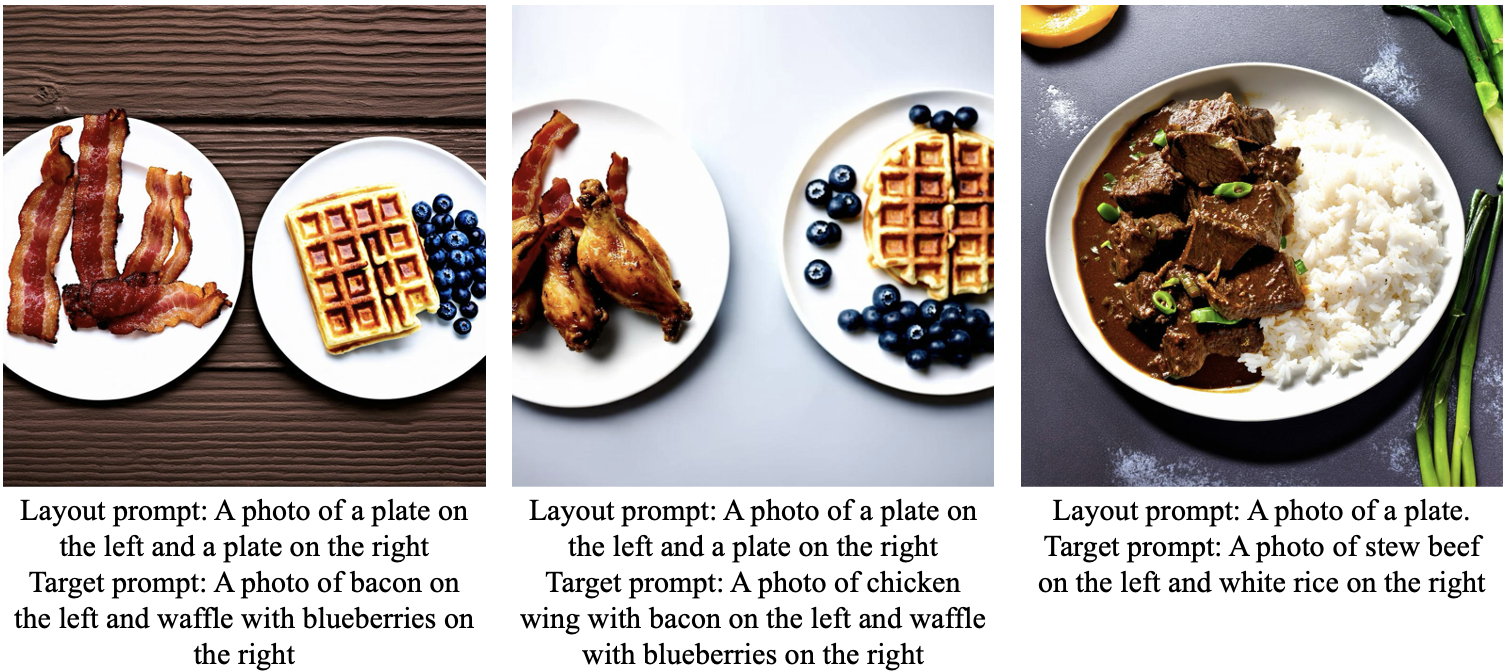}
   \caption{Controllable Food Image Generation}
   \label{fig:con_food}
\vspace{-2mm}
\end{figure}

Not all food items need to be separated. Users sometimes prefer intentional co-location (e.g., stew beef with rice). Our framework supports this controllability by specifying fewer layout regions during the early layout phase. For example, in Figure \ref{fig:con_food}, generating a single plate in the layout step places stew beef and rice together. Thus, beyond enforcing separation, PG can intentionally entangle items when desired, simply by adjusting the number of layouts specified in the prompts.

\subsection{Generated Image Results in Non-Food Domain}

\begin{figure}[t]
   \centering
   \includegraphics[width=0.95\linewidth]{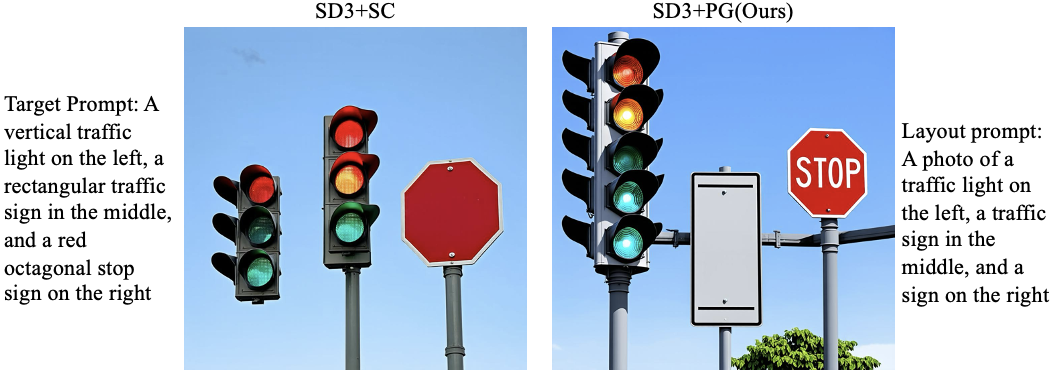}
   \caption{Generated image results in non-food domain}
   \label{fig:nonfood}
\vspace{-3mm}
\end{figure}

To demonstrate that our PG  generalizes beyond the food domain, we conducted qualitative experiments on three representative non-food object pairs, as shown in Figure \ref{fig:nonfood}. Although in non-food domain, it generally does not have object entanglement problem, our method can also address other types of issue. In the example shown, the target prompt is complex. However, our method can simplify the prompt by firstly focusing on the core part in the prompt during layout steps and then focus on other details in the target prompt so that the generated image at least has correct layout without confusion of placement objects in the image. 

\vspace{-1mm}

\section{Conclusion}
\label{sec:concl}


In this paper, We identified and addressed entanglement
problem in compositional food generation via
PG, showing consistent gains in recall and object existence
across two datasets. Future work will explore automatic spatial relation inference.

\vspace{-1mm}

\bibliographystyle{IEEEtran}
\small \bibliography{egbib}

\end{document}